\documentclass[conference]{IEEEtran}
\IEEEoverridecommandlockouts

\usepackage{cite}
\usepackage{amsmath,amssymb,amsfonts}
\usepackage{algorithmic}
\usepackage{graphicx}
\usepackage{textcomp}
\usepackage{xcolor}
\usepackage{times}
\usepackage{soul}
\usepackage{url}
\usepackage[hidelinks]{hyperref}
\usepackage[utf8]{inputenc}
\usepackage[small]{caption}
\usepackage{booktabs}
\usepackage{latexsym}
\usepackage{amsmath,amssymb,amsfonts,amsthm}
\usepackage{booktabs}
\usepackage{algorithm}
\usepackage{float}
\usepackage{multirow}
\hypersetup{draft}
\usepackage{threeparttable}

\def\BibTeX{{\rm B\kern-.05em{\sc i\kern-.025em b}\kern-.08em
    T\kern-.1667em\lower.7ex\hbox{E}\kern-.125emX}}

\begin{document}

\title{Adversarially Robust and Explainable Model Compression with On-Device Personalization for Text Classification}

\author{\IEEEauthorblockN{Yao Qiang, Supriya Tumkur Suresh Kumar, Marco Brocanelli and Dongxiao Zhu}
	    \IEEEauthorblockA{\textit{Department of Computer Science}} 
		\textit{Wayne State University} \\ 
		Detroit, MI, USA \\
		Email: \{yao, supriyats, brok, dzhu\}@wayne.edu}

\maketitle

\begin{abstract}

On-device Deep Neural Networks (DNNs) have gained more attention due to the increasing computing power of mobile devices and applications in Computer Vision (CV), Natural Language Processing (NLP), and Internet of Things (IoTs) recently. Unfortunately, the existing efficient Convolutional Neural Networks (CNNs) initially designed for CV tasks are not directly applicable to NLP tasks, and most tiny Recurrent Neural Networks (RNNs) are applied primarily for IoT applications. Although several model compression techniques have seen initial success in multiple on-device text classification tasks, there are at least three major challenges yet to be addressed: adversarial robustness, explainability, and personalization in NLP applications. We attempt to tackle these challenges by designing a new training scheme, which builds the adversarial robustness and explainability in our compressed RNN model during the training process via simultaneously optimizing the adversarially robust objective and the explainable feature mapping objective. The resulting compressed model is personalized using on-device private training data via fine-tuning. We perform extensive experiments demonstrating the effectiveness of our approach by comparing our compressed models with both compact RNNs (e.g., FastGRNN) and compressed RNNs (e.g., PRADO) in both clean and adversarial test settings. 
	
\end{abstract}

\section{Introduction}

As the improved hardware and increased availability of mobile apps are expected to provide personalized user experience, protect users' privacy, and offer minimum latency, mobile Artificial Intelligence (AI) finds application in a wide range of domains, including image classification, healthcare, and speech recognition recent years. Even with the rapid improvements of mobile hardware, the main obstacles in deploying the standard DNNs on mobile devices are the high computational and memory requirements during both training and inference processes. Recently, several model compression techniques have been applied to develop compact DNNs for mobile deployment \cite{rastegari2016xnor,han2015deep}. An array of lightweight CNN architectures has been designed specially for CV tasks \cite{sandler2018mobilenetv2,howard2019searching}. In addition, the compact RNN architectures are mostly developed and applied in lightweight (IoTs) applications, to learn from on-device sequential/temporal data. Despite the initial success, the development of on-device NLP applications has lagged behind applications in other domains. Only a small number of compressed RNN models are available for resource-hungry NLP applications \cite{ravi2017projectionnet}. 

Up-to-date, the vast majority of mobile AI applications focus on inference since the more computationally intense on-device training has not been demonstrated as a feasible task. In addition to the computational challenge, a trustworthy compressed RNN model needs to tackle the following major challenges: (1) information loss in model compression, (2) adversarial robustness, and (3) lack of explainability and personalization. Systematically addressing these challenges would significantly promote the wide adoption of the compressed RNN models in NLP applications.

In this work, we design a novel training scheme (Figure \ref{fig:overview}) to address the above challenges via building the adversarial robustness and explainability in our compressed RNN model during the training process, followed by on-device personalization. Thus, our approach avoids the post-hoc adversarial training and attribution based explanation. Although we only demonstrate impressive performance against several adversarial attacks towards text classification tasks, our training scheme is applicable to other RNN-based NLP tasks.

Specifically, our major contributions are: (1) we tackle the information loss in model compression by minimizing the layer-wise feature mapping loss and label distillation loss during the training process, (2) we employ an adversarial robust objective function to enhance model robustness via encouraging the predicted probabilities of false classes to be equally distributed in both the full and compressed models, (3) we enable the model explainability via an explainable aspect-based feature mapping. Furthermore, we retain model personalization via fine-tuning the on-device compressed model using the on-device training data, and (4) our extensive experiments demonstrate the effectiveness of our approach by comparing with baseline compact and compressed RNNs in multiple text classification tasks. 

%\begin{itemize}
%	\item We tackle the information loss in model compression by minimizing the layer-wise feature mapping loss and label distillation loss during the training process.
%	\item We employ an adversarial robust objective function to enhance model robustness via encouraging the predicted probabilities of false classes to be equally distributed in both the full and compressed models.
%	\item We enable the model explainability via an explainable aspect-based feature mapping. Furthermore, we retain model personalization via fine-tuning the on-device compressed model using the on-device training data. 	
%	\item Our extensive experiments demonstrate the effectiveness of our approach by comparing with baseline compact and compressed RNNs in multiple text classification tasks. 
%\end{itemize}	

\section{Related Work}

\subsection{Adversarial Attacks and Defense}

Both white-box and black-box adversarial attacks \cite{zhang2020adversarial} generate crafted adversarial examples to fool DNNs into making wrong predictions. In the former, the attacks have complete access to the target model (e.g., model architecture and parameters) and generate adversarial samples exploiting the guidance of the model's gradients \cite{goodfellow2014explaining}. In the latter, the attacks are only allowed to test the target classifiers without any access to the target model \cite{liang2017deep}. We select three strong attack methods \cite{samanta2017towards,gao2018black,ren2019generating} covering both black-box and white-box attacks, as will be described in the Experiment Setup section.

Recently, an array of defense techniques has been proposed to counter these adversarial attacks. As an alternative to the traditional \textit{natural training}, which is vulnerable to carefully crafted attacks, \textit{adversarial training} is a technique that allows DNNs to correctly classify both clean and adversarial examples \cite{goodfellow2014explaining}. However, most existing adversarial training approaches are based on one specific type of adversarial attacks, leading to a compromised generalization of defense capability on adversarial examples from other attacks \cite{song2018improving}. Besides, the high computing cost in generating strong adversarial examples makes the standard adversarial training computationally prohibitive \cite{shafahi2019adversarial}, especially on large-scale NLP datasets. 

In tackling these problems, efficient approaches have been proposed \cite{pang2019rethinking} to improve the adversarial robustness of DNNs via learning discriminative features through minimizing well-designed loss functions. \cite{chen2019complement} proposed a novel loss objective called Complement Objective Training (COT) that achieved good robustness against single-step adversarial attacks via maximizing the likelihood of the ground-truth classes while neutralizing the probabilities of the complementary (false) classes using two different loss objectives. However, these two loss objectives did not have a coordination mechanism to work together efficiently. In order to reconcile the competition between them, a new approach called Guided Complement Entropy (GCE) has been recently proposed \cite{chen2019improving} for CV applications. Specifically, GCE adds a ``guided" term to maintain the balance between the true class and the false classes, which helps improve the adversarial robustness. Without loss of generality, we employ GCE to extract adversarially robust features from large-scale NLP datasets and demonstrate the improvement of the adversarial robustness of our models against various adversarial attacks. To the best of our knowledge, we are among the first to leverage a well-designed loss function to enable the adversarial robustness of model compression for NLP applications.

\subsection{Compressed RNN Models}

Several model compression techniques have been demonstrated to achieve compactness, e.g., designing tailor-made lightweight architectures \cite{sandler2018mobilenetv2} and generating compressed models via automatic Neural Architecture Search \cite{tan2019mnasnet}. Quantization \cite{rastegari2016xnor} and network pruning \cite{han2015deep} reduce the model size at the expense of reduced prediction accuracy. Knowledge distillation \cite{hinton2015distilling} is used to improve the performance of the lightweight models by transferring the knowledge from a large teacher network to a lightweight student network.

In order to deal with on-device text classification tasks, \cite{ravi2017projectionnet} proposed a new architecture that jointly trains a full neural network and a simpler projection network leveraging random projections to transform inputs or intermediate representations into bits. The projection network encoded lightweight and efficient computing operations in bit space to reduce memory footprint. Since then, a few more advanced projection networks have been proposed to achieve better performance. Self-Governing Neural Network (SGNN) \cite{ravi2018self} was proposed to learn compact projection vectors with local sensitive hashing, which surmounted the need for pre-trained word embeddings with huge parameters in complex networks. \cite{krishnamoorthi2019prado} proposed a novel projection attention neural network named PRADO combing trainable projections with attention and convolutions. The compressed 200-Kilobytes model had achieved good performance on multiple text classification tasks. 

However, most of the above approaches overlook the important issue of adversarial robustness in NLP applications. Recent literature has focused on the trade-off between adversarial robustness and neural network compression with application in CV. \cite{zhao2018compress} investigated the extent to which adversarial samples are transferable between uncompressed and compressed DNNs. \cite{gui2019model} proposed a novel Adversarially Trained Model Compression (ATMC) framework that obtains a remarkably favorable trade-off among model size, accuracy, and robustness. \cite{ye2019adversarial} proposed a framework of concurrent adversarial training and weight pruning that enables model compression while still preserving the adversarial robustness. Different from all the above state-of-the-art solutions, we tailor-made model compression training scheme for NLP to enable adversarially robust, explainable, and personalized on-device applications.   

\begin{figure*}[ht]
	\centering 
	\includegraphics[width=0.80\linewidth]{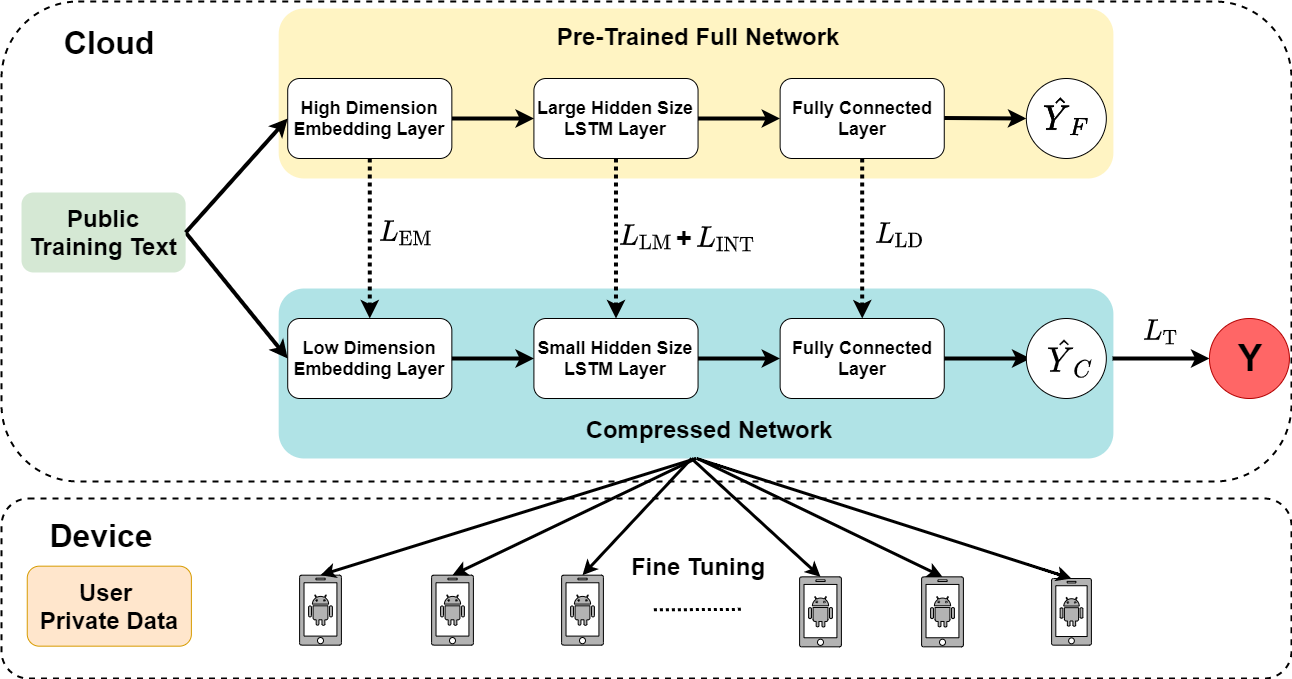}
	\caption{Our training scheme for model compression. The full model is pre-trained on the cloud. The compressed model is trained on the cloud with our training scheme using public datasets to ensure adversarial robustness and explainability, followed by fine tuning and deploying on mobile devices with on-device private data.}
	\label{fig:overview}
\end{figure*}

\subsection{Compact RNN Models}

RNN models have achieved significant success in learning complex patterns for temporal/sequential data (e.g., sensor signal, natural language). Beyond the classical LSTM and GRU architectures, more sophisticated RNNs with skip connections and residual blocks \cite{campos2017skip} and those combined with CNNs have been developed to allow the RNNs to go `deeper' and achieve better performance. Despite the state-of-the-art performance, these heavyweight RNN models are resource-hungry and not suitable for on-device deployment.  

More recently, tiny RNNs with small parameter sizes (e.g., 10K or less), have received increasing attention due to their high application potential to on-device deployment in IoT environment. \cite{kusupati2018fastgrnn} proposed FastRNN/FastGRNN by adding residual connections and gating on the standard RNNs, which outperformed LSTM and GRU in prediction accuracy with fewer parameters (10K versus 30K). AntisymmetricRNN \cite{chang2019antisymmetricrnn} is designed based on ordinary differential equations (ODEs); this model achieves comparable performance with both LSTM and GRU with a much smaller number of parameters (10K). iRNN \cite{kag2019rnns} is a similar model based on ODEs, which is designed to facilitate the RNN training with identity gradient. It achieves a better performance than GRU and a comparable performance to LSTM with only 7.80K model parameters. Despite the impressive performance in IoT applications, thus far only FastGRNN \cite{kusupati2018fastgrnn} has demonstrated a pilot NLP application with a much larger model size (250K) compared to the lightweight IoT applications (10K).   

\section{Method}

\subsection{Training Scheme}

Figure \ref{fig:overview} illustrates our training scheme, which is composed of a new compression technique to minimize information loss, aspect-based feature mapping to ensure explainability, and on-device fine-tuning to enable personalization. Specifically, we design feature mapping and label distillation layers to minimize information loss during model compression. Additionally, the aspect-based feature mapping method enables the model explainability via minimizing the interpretable loss during training. Finally, the compressed model is fine-tuned with different values of hyper-parameter $T$ in the label distillation layer, leveraging on-device training data for personalization.

We denote an input text as $\{w_1, \dots, w_i, \dots, w_n\}$, where $i$ is the index and $n$ is the number of words. The first layer in most models designed for NLP tasks applies an embedding layer with trainable parameters $\mathbf {W}_e \in \mathbb{R}^{d \times V}$ to map each word $w_i$ to a fixed-length $d$-dimension vector $\mathbf{e}_i \in \mathbb{R}^d$, where $V$ denotes the vocabulary size. The embedded word vectors $\mathbf e$ are then processed by the remaining layers. To retain sufficient embedding information, most models use a large vocabulary size $V$ (ranging between hundreds of thousands to millions) and a high embedding dimension $d$ (e.g., 100 or higher), leading to a huge number of parameters in $\mathbf {W}_e$. As there is a minimum required vocabulary size to achieve a specific performance \cite{chen2019large}, we set the embedding layer to a low dimension (e.g., 5, 10, and 20) in the compressed model to decrease the number of parameters in $\mathbf {W}_e$. In order to minimize the embedding information loss, we design a feature mapping method that allows the compressed model to learn embedding information from the pre-trained full model by minimizing the difference between the two embedded features. Specifically, we first employ an autoencoder to compress the high dimension embedding feature $\mathbf e_f$ in the full model into the low dimension $\mathbf{\hat{e}}_f$ embedding feature. We then minimize the distance $||\mathbf{\hat{e}}_f - \mathbf e_c||_2$ ensuring $\mathbf{\hat{e}}_f \approx \mathbf e_c$ during training. $f$ and $c$ indicate the full and compressed models, respectively. 

As shown in Figure \ref{fig:overview}, following the embedding layer is a recurrent neural network layer, e.g., Long Short-Term Memory (LSTM). The number of parameters in LSTM layer can be calculated as: $4 \times H \times ((d + 1) + H)$ \cite{kuchaiev2017factorization}, where $H$ is the size of latent feature and $d$ is the input size. So the latent feature size $H$ can greatly affect the number of parameters. As such, we set a small hidden state size in LSTM hidden layer without using attentions, short-cut connections, or other sophisticated additions to reduce the number of parameters in the compressed model. We employ another autoencoder to encode the high dimension latent features $\mathbf h_f$ into the low dimension $\mathbf{\hat{h}}_f$. Then we minimize the distance $||\mathbf{\hat{h}}_f - \mathbf h_c||_2$ to ensure $\mathbf{\hat{h}}_f \approx \mathbf h_c$. Additionally, we design an aspect-based feature mapping in this layer to enable the explainability of the model, which is explained in more detail in the next section. 

The third layer in our training scheme exploits label distillation, which enforces the compressed model to mimic the prediction behavior of the full model by training the former with more informative soft labels generated by the latter \cite{hinton2015distilling}. Furthermore, \cite{liang2017enhancing} has demonstrated that a good manipulation of temperature $T$ can push the softmax scores of in- and out-of-distribution (OOD) images further apart from each other, making the OOD images more distinguishable. Inspired by this, we fine-tune the compressed model on each user's device with personalized training data with different values of hyper-parameter $T$ to achieve on-device personalization. 

\subsection{Explainable Feature Mapping}
\label{sec:explain} 

\begin{figure}[t]
	\centering
	\includegraphics[width=0.75\linewidth]{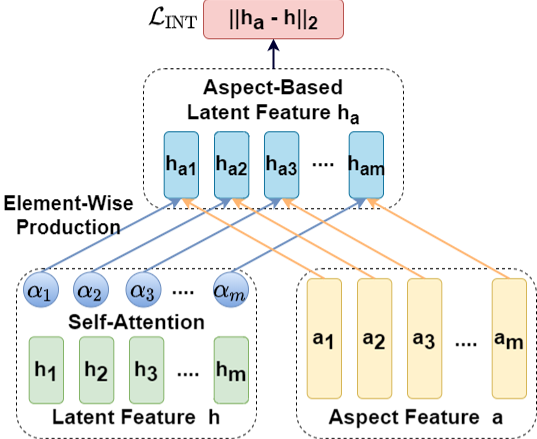}
	\caption{Explainable feature mapping.}
	\label{fig:projection}
\end{figure} 

Our aspect-based feature mapping (Figure \ref{fig:projection}) enables the explainability of the model compression leveraging the ubiquitous aspect information, which is an additional input feature of our models. Aspect-based explainable models have been developed to solve problems in collaborative filtering and sentiment analysis \cite{qiang2020tagfree}. For example, \textit{food}, \textit{price}, \textit{service}, and \textit{ambience} are aspects in Restaurant domain to explain the sentiment polarities of the user reviews, and \textit{genres} are used as aspects in Movie domain for recommendation. We first apply one-hot encoding to represent a particular aspect domain, then embed these aspect one-hot vectors into an aspect-guided latent space with the same embedding dimension $d$ in the embedding layer, denoted as $\mathbf{a} = \{\mathbf{a}_1, \ldots, \mathbf{a}_j, \ldots, \mathbf{a}_m\}$, where $j$ is the index and $m$ is the number of the aspects. Then, we map the latent features $\mathbf h$ from the LSTM layer into the same space with these aspect features, and get the aspect attention weights through a self-attention mechanism, where $\alpha_j  = \text{Softmax}(\mathbf h)$ is the attention weight for $j$-th aspect. In this way, we derive a new aspect-based latent features $\mathbf {h}_a$ by combining the aspect features $\mathbf{a}$ with the aspect attention weights $\boldsymbol{\alpha}$ as: $\mathbf {h}_a = \sum_{j=1}^{m} \mathbf {\alpha}_j \mathbf{a}_j$. Then, $\mathbf {h}_a$ is passed to the following fully connected layers. Our goal here is to minimize the distances between the general feature and the aspect-based features (i.e., $||\mathbf{h}_a - \mathbf{h}||_2$) to ensure the similarity between them (i.e., $\mathbf h \approx \mathbf{h}_a$). Therefore, our approach makes the model explanation intrinsically and can really answer ``how" the compressed model generates corresponding predictions supervised by the novel interpretable loss objective \cite{pan2020explainable}. In the experiments, we employ a novel metric (hit-ratio) as a quantitative evaluation of our model intrinsic explainability, as will be described in the Results and Discussion section.

\subsection{Training Objective}

We select different loss functions for the components in our training scheme according to the different training objectives. We employ a well-designed loss function GCE \cite{chen2019improving} instead of the traditional CrossEntropy (CE) not only to improve the adversarial robustness of the full model and the compressed model but also to maximize the transfer of robustness from the full model to the compressed model. Thus, the text classification task loss is denoted $\mathcal{L}_{\mathrm{T}}$ = GCE($y,\hat y$). We minimize the Mean Square Errors (MSE) between the high-dimension features from the full model and the low-dimension features from the compressed model aiming to assimilate them in the two feature mapping layers: $\mathcal{L}_{\mathrm{EM}}$ = $\mathrm{MSE}(\mathbf{\hat{e}}_f,\mathbf e_c)$ and $\mathcal{L}_{\mathrm{LM}}$ = $\mathrm{MSE}(\mathbf{\hat{h}}_f,\mathbf h_c)$, respectively. Additionally, we minimize the Kullback–Leibler (KL) divergence loss, denoted as $\mathcal{L}_{\mathrm{LD}}$ = $\mathrm{KLDiv}(\text{Softmax}(\mathbf o_f/T),\text{Softmax}(\mathbf o_c/T))$, between the output label probability distributions in the label distillation layer. For the two autoencoders used in the feature mapping layers, we minimize MSE between the inputs and the reconstructed outputs to derive a better feature mapping: $\mathcal{L}_{\mathrm{AE}}$ = $\mathrm{MSE}(\mathbf e_f,\mathbf e_r)$ + $\mathrm{MSE}(\mathbf h_f,\mathbf h_r)$, $\mathbf e_r$ and $\mathbf h_r$ here are the reconstructed outputs. We also use MSE as the model interpretable loss: $\mathcal{L}_{\mathrm{INT}}$ = $\mathrm{MSE}(\mathbf h_a,\mathbf h)$ to derive explainable feature mapping. 

The general training objective function consists of several components, formally: $\mathcal{L}$ = $\lambda_1\mathcal{L}_{\mathrm{T}}$ + $\lambda_2 (\mathcal{L}_{\mathrm{EM}}$ + $\mathcal{L}_{\mathrm{LM}}$ + $\mathcal{L}_{\mathrm{LD}})$ + $\lambda_3 \mathcal{L}_{\mathrm{AE}}$ + $\lambda_4 \mathcal{L}_{\mathrm{INT}}$, where $\lambda_1$, $\lambda_2$, $\lambda_3$, and $\lambda_4$ are tuning hyper-parameters to leverage the relative importance of different loss objectives. With that, the compressed model is duly regularized by the pre-trained full model. We note {\bf the general loss objective is sufficiently flexible} that each loss objective can be retained/dropped according to the real-world need. 

\section{Experiment Setup}

Our experiments are conducted with multiple text classification tasks, such as sentiment analysis (Amazon and Yelp), news categorization (AG's News), and topic classification (Yahoo). In particular, we want to answer the following questions through the experiments: (1) Can the compressed model achieve a competitive performance with the cloud-based model through our new training scheme? (2) Does the compressed model ensure adversarial robustness against strong adversarial attacks? (3) Is the compressed model explainable for on-device deployment?

\subsection{Datasets}

Statistics of the text classification datasets used in our experiments are listed in Table \ref{Datasets Statistics}. We conduct our explainable feature mapping experiments on Rest-2014 dataset, which is from SemEval 2014 \cite{pontiki-etal-2014-semeval} containing reviews of restaurant domains together with some aspect features such as \textit{food}, \textit{price}, \textit{service}, and \textit{ambience}. We select Senti140 from LEAF \cite{caldas2018leaf} to evaluate the performance of our on-cloud model compression and on-device personalization. Specifically, we randomly select 5,000 different users from the Senti140 dataset, including 18,400 positive samples and 16,200 negative samples, to train the on-cloud model and compress the on-device model. After this, we select the user devices containing the most abundant training samples to fine-tune the compressed models for further improving the performance and retaining model personality. 

\begin{table}[ht]
	\centering 
	\resizebox{0.40\textwidth}{!}{
		\begin{tabular}{l|r r r}
			\hline
			\textbf{Dataset}      & \textbf{\#Training} & \textbf{\#Testing} & \textbf{\#Class} \\ \hline
			Amazon          	  & 3,600,000  & 400,000   & 5         \\ 
			\hline
			Yelp            	  & 560,000    & 38,000    & 5         \\ 
			\hline
			AG's News       	  & 120,000    & 7,600     & 4         \\ 
			\hline
			Yahoo           	  & 1,400,000  & 60,000    & 10        \\ 
			\hline
			Rest-2014       	  & 3,000      & 800       & 3         \\
			\hline
		\end{tabular}
	}
	\caption{Datasets Statistics Details}
	\label{Datasets Statistics}
\end{table}

\begin{table*}[h]
	\centering 
	\resizebox{\textwidth}{!}{
		\begin{tabular}{l|l|c|c|c|c|c|c|c|c}
			\hline
			\multirow{3}*{\textbf{Dataset}}  &\multirow{3}*{\textbf{Loss}}  &\multicolumn{4}{c}{\textbf{Conventional Training Scheme}}       & \multicolumn{4}{c}{\textbf{Our Training Scheme}}   \\ 
			\cline{3-10}
			&								&\textbf{Clean}&\textbf{PWWS}&\textbf{Gradient}&\textbf{Replaceone} &\textbf{Clean}&\textbf{PWWS}&\textbf{Gradient}&\textbf{Replaceone}   \\
			%\cline{3-10}
			&&Acc/F1		&AdvAcc/AdvF1	&AdvAcc/AdvF1	 &AdvAcc/AdvF1			&Acc/F1     	&AdvAcc/AdvF1  &AdvAcc/AdvF1	&AdvAcc/AdvF1   \\
			\hline 
			\multirow{2}*{Amazon}	&CE		& 58.8/45.2  & 30.5/25.4	& 49.2/34.8    & 37.6/31.0  		& 59.0/45.3	& \textbf{33.1}/24.9  & 50.1/39.8	& 39.4/35.0  	   \\ 
			
			&GCE	& 58.0/44.1  & 32.1/24.8	& 50.1/37.7    & 37.9/29.4  		& \textbf{59.8/45.7}	& 32.4/\textbf{26.1}  & \textbf{58.7/44.6}	& \textbf{49.0/38.4}  	   \\ 
			\hline 
			\hline 
			%\hline 
			\multirow{2}*{Yelp}	    &CE		& 61.0/45.9  & 36.4/30.1	& 43.7/30.1    & 42.4/29.4  		& \textbf{62.7/47.7}	& \textbf{36.6/29.3}  & 56.4/46.8	& 43.8/30.4  	   \\ 
			
			&GCE	& 59.8/45.4  &34.2/30.6 	& 56.8/40.1    & 43.7/30.1  		& 61.6/46.7	        & 35.4/28.1  & \textbf{62.5/48.0}	& \textbf{45.3/32.2} 	   \\ 
			\hline 
			\hline 
			%\hline 
			\multirow{2}*{Yahoo}	&CE		& 71.5/65.2  & 40.6/35.3	& 56.6/51.1    & 43.8/42.7 		    & \textbf{72.2/65.9}	& 42.1/38.9  & 58.1/53.6	& 45.8/43.9  	   \\ 
			
			&GCE	& 71.2/65.0  & 39.6/36.5	& 61.5/57.3    & 45.7/43.4 		    & 71.5/65.1	        & \textbf{43.7/40.4}  & \textbf{66.6/64.6}	& \textbf{49.6/48.3}  	   \\ 
			\hline 
			\hline 
			%\hline 
			\multirow{2}*{AG's NEWS}&CE		& 89.5/67.0  & 52.4/46.9	& 88.2/66.2    & 68.2/51.4          & 91.0/68.3	& 55.4/47.1  & 89.6/\textbf{69.1}	& 70.3/55.3 	   \\ 
			
			&GCE	& 90.5/67.9  & 56.7/48.9	& 89.3/68.3    & 74.4/55.8 		& \textbf{91.1/68.4}	& \textbf{58.2/49.0}  & \textbf{90.2}/67.4	& \textbf{79.6/62.2} 	   \\ 
			\hline  
		\end{tabular}
	}
	\caption{Comparison of {\bf performance} and {\bf adversarial robustness} of the compressed models using the conventional and our training schemes on four benchmark datasets. Conventional training scheme represents the traditional training process of RNN networks, which we used to train our full model. Clean here indicates natural test samples. PWWS, Gradient, and Replaceone are three different adversarial attacks. Best performance are bold-faced.} 
	\label{main} 	
\end{table*}

\subsection{Adversarial Attacks}
\label{sec:attacks} 

Most strong adversarial attacks \cite{zhang2020adversarial} target on the pre-trained embedding matrix (e.g., \cite{ebrahimi2017hotflip}). Others (e.g., \cite{zhang2019generating}) need to apply additional meta-data such as SentiWordNet \cite{baccianella2010sentiwordnet}. Both methods iteratively search through the embedding matrix requiring a long time to find the effective perturbations, representing a substantial obstacle for attacking in real-time. Furthermore, their performance on the transferred attacks decrease dramatically compared to that on the original architecture and data, demonstrating a poor generalization ability \cite{zhang2020adversarial}. Therefore, we select two strong one-off attacks, i.e., Replaceone \cite{gao2018black} and Gradient \cite{samanta2017towards}, which are more efficient and with the minimal alteration that is more suitable for real-world applications. We also choose another strong attack, i.e., probability weighted word saliency (PWWS) \cite{ren2019generating}, which makes the perturbations not only unperceivable but also preserves maximal semantic similarity. We use the same settings as in \cite{gao2018black,ren2019generating} for our experiments. In more detail:  
\\
{\bfseries Replaceone.} \cite{gao2018black} applied several different black-box scoring functions and word transformation methods to generate adversarial samples with minimum word changes. We use Replaceone, a more efficient yet effective scoring function, to find the most important words then swap two adjacent letters to generate new words in the adversarial samples.   
\\
{\bfseries Gradient.} \cite{samanta2017towards} proposed a white-box attack that first used gradients to identify salient words in the original samples and then modified these words to generate adversarial samples. 
\\
{\bfseries PWWS.} \cite{ren2019generating} proposed a greedy algorithm based white-box attack using a new word replacement order determined by both the word saliency and the classification probability to generate adversarial samples with lexical/grammatical correctness and semantic similarity.

% We generate 1000 adversarial samples using PWWS then evaluate our models' robustness with these adversarial samples together with our test sets (Did you use the same number for other attacks, if yes, move it). 
% alzantot2018generating
\subsection{The Compared Methods}

Our goal is to develop adversarially robust model compression for NLP tasks, which not only achieves competitive performance in classifying both clean and adversarial samples but also satisfies the on-device resource constraints. Although our compressed model is intended for on-device deployment, we compare with some on-cloud baseline models that are designed to exploit the full extent of cloud resources, i.e., bag-of-words TF-IDF, N-grams TF-IDF, char-level CNN, word-level CNN, and LSTM \cite{zhang2015character} to demonstrate our competencies. We also compare our compressed models with other compressed and compact models designed for on-device applications. We compare with two existing on-device compressed neural network models for NLP applications (i.e., SGNN \cite{ravi2018self} and PRADO \cite{krishnamoorthi2019prado}) in terms of the test performance using clean samples only since (1) they are not designed for defending against adversarial attacks and (2) there are no source codes available. Moreover, we compare with the compact model, FastGRNN \cite{kusupati2018fastgrnn}, which is originally designed for IoT applications and recently demonstrated promising performance in NLP applications. We discuss the comparison results in Table \ref{baseline}. We provide more implementation details and our hyper-parameters tuning in Table \ref{Hyper}. We will release our code in future proceedings. 

\begin{table} [hbt!]
	\centering 
	\resizebox{0.40\textwidth}{!}{
		\begin{tabular}{c|c|c}  
			\hline  
			\textbf{Hyper-parameters}&\textbf{Values}&\textbf{Optimal Selection}\\
			\hline
			Learning Rate&0.01; 0.005; 0.001&0.001\\
			\hline
			Batch size&32; 64; 128&64\\
			\hline
			$T$&20, 50, 80, 100&80\\
			\hline
			$\lambda_1$&0; 0.2; 0.5&0.2\\
			\hline
			$\lambda_2$&0.2; 0.5; 0.8&0.8\\
			\hline
			$\lambda_3$, $\lambda_4$&0.2; 0.5; 0.8&0.5\\
			\hline
		\end{tabular}
	}
	\caption{Hyper-parameters tuning}  
	\label{Hyper} 	
\end{table}

\subsection{Evaluation Metrics}

In our experiments, we not only consider the performance of the compressed models trained with clean texts but also their adversarial robustness towards various adversarial attacks. We evaluate the models' performance in terms of accuracy (Acc) and macro-F1 scores (F1) for text classification tasks. We also test our model robustness on thousands of adversarial examples generated from different adversarial attacks separately. We report the adversarial accuracy (AdvAcc) and adversarial macro-F1 score (AdvF1) to evaluate the adversarial robustness of the models. In addition, we design a novel hit-ratio to evaluate our model explainability. 

\section{Results and Discussion}

\subsection{Performance Comparison}
\label{sec:main} 

Table \ref{main} shows the performance and adversarial robustness of the compressed models, either trained in a conventional way or in the proposed training scheme (ours). It is observed that our compressed models achieve an overall better performance on all the evaluation metrics. As our new training scheme exploits feature mapping and label distillation to enable the compressed models to learn information from the pre-trained full models, it is capable of reducing the information loss during model compression, leading to marked improvement on clean sample accuracy and macro-F1 scores. In addition, the compressed models trained with GCE loss improve the adversarial robustness compared to the ones trained with the CE loss in a vast majority of comparisons, highlighting the advantage of adversarially robust model compression. 

\begin{table*}[h]
	\centering
	\resizebox{\textwidth}{!}{
		\begin{threeparttable}
			\begin{tabular}{l|l|c|c|c|c|c|c|c|c|c}
				\hline
				\multirow{3}*{\textbf{Type}}	&\multirow{3}*{\textbf{Method}}  			 &\multicolumn{3}{c|}{\textbf{Yelp}}                & \multicolumn{3}{c|}{\textbf{Amazon}}                &\multicolumn{3}{c}{\textbf{Yahoo}}                \\ 
				\cline{3-11}
				&&\textbf{Clean}&\textbf{Gradient}&\textbf{Replaceone}     &\textbf{Clean}&\textbf{Gradient}&\textbf{Replaceone}  &\textbf{Clean}&\textbf{Gradient}&\textbf{Replaceone} \\
				&&Acc  &AdvAcc&AdvAcc      &Acc&AdvAcc&AdvAcc 			 &Acc&AdvAcc&AdvAcc \\
				\hline
				\multirow{3}*{Compressed}& {\bf Ours} & 61.7&\textbf{62.5}&\textbf{45.4}   & 59.9& \textbf{58.8}&49.0     & 71.5& \textbf{66.6}&\textbf{49.6}  \\
				&PRADO& \textbf{64.7}$^\star$& N/A&N/A   		  & \textbf{61.2}$^\star$& N/A&N/A     		  & \textbf{72.3}$^\star$& N/A&N/A  		 \\ 
				&SGNN& 35.4$^\star$& N/A&N/A   		  & 39.1$^\star$& N/A&N/A      & 36.6$^\star$& N/A&N/A           \\ 
				\hline
				\hline
				Compact	& FastGRNN    & 26.73& 20.74&20.68   		  & 30.20& 20.41&19.66     & 28.34& 19.74&21.67           \\ 
				\hline
				\hline                               		 
				\multirow{6}*{Full}	& {\bf Ours}      & 62.4& 56.7&45.0   & 60.2& 53.5&\textbf{53.3}     & 72.1& 64.7&45.8  \\ 
				& CNN-char                  & 62.0& 45.7&40.8   & 59.6& 47.0&42.1     & 71.2& 43.5&44.9  \\ 
				& CNN-word                  & 60.5& 44.8&37.6   & 57.6& 47.6&41.1     & 71.2& 51.9&46.6  \\ 
				& LSTM                      & 58.2& 53.2&42.6   & 59.4& 55.7&41.3     & 70.8& 50.0&45.1  \\ 
				& BoW TFIDF                 & 59.9& N/A&N/A   & 55.3& N/A&N/A     & 71.0& N/A&N/A  \\ 
				& N-gram TFIDF              & 54.8& N/A&N/A   & 52.4& N/A&N/A     & 68.5& N/A&N/A  \\ \hline
			\end{tabular}
			\begin{tablenotes}
				\item[1] N/A: not applicable. $^\star$: results reported in \cite{krishnamoorthi2019prado}.
			\end{tablenotes}
		\end{threeparttable}
	}
	\caption{Comparison of our compressed models with other compressed, compact and full models. Our compressed and full models have 200K and 2M parameters. PRADO has 175 K parameters and FastGRNN has more than 250K parameters.} %Best performance are bold-faced.}
	\label{baseline} 
\end{table*}

We further compare the performance of our compressed models with the other compressed, compact models and full models in Table \ref{baseline} in both clean and adversarial example settings (wherever applicable). Note the results of PRADO \cite{krishnamoorthi2019prado} and SGNN \cite{ravi2018self} are cited directly from the original papers since no source codes are made available. Therefore, we only compare the clean sample accuracy with these two compressed RNN models since they are not designed for mitigating adversarial attacks and hence not adversarially robust. In Table \ref{baseline}, our compressed model achieves a better performance than SGNN. Moreover, it maintains a comparable clean sample accuracy with PRADO yet demonstrates impressive adversarial robustness, which is one of the unique features of our model compression technique. 

We also compare our compressed RNN model with a manually designed compact RNN model named FastGRNN \cite{kusupati2018fastgrnn}. Specifically, we re-implemented the algorithm based on the package released in the original paper (\url{https://pypi.org/project/edgeml-pytorch/}). We keep the same settings of adversarial attack methods to make a fair comparison in terms of adversarial robustness between our compressed model and the compact FastGRNN. From Table \ref{baseline} we observe better performance of our compressed model on both clean and adversarial examples than FastGRNN. Both clean and adversarial accuracies are critical for on-device NLP applications. For example, in cybersecurity, detection of business email compromise (BEC) and email account compromise (EAC) scams has become a rising challenge since it does not contain malware but only social engineering messages. As such, the on-device NLP app needs to look into emails' text contents to perform the adversarial attack detection. Finally, we show that our models outperform a number of competing full models in terms of both clean and adversarial accuracies.

\subsection{Model Explainability}
In Method section, we describe our explainable aspect-based feature mapping as an optional component of model compression training scheme. We first calculate attention weights $\mathrm{score}_j$ for $j$-th aspect by adding them up for each token in texts, formally: $\mathrm{score}_j$ = $\sum_{i=1}^{n} \mathbf {\alpha_{ji}}$. Then we identify the aspect as the one with the highest attention weight denoted as $A_H$, and compare with the ground truth aspect coded in the one-hot vector $A$. Note $A_H$ and $A$ here are the index values. If these two are consistent, formally: $ A_H = A $, we consider it as a successful hit, demonstrating that the model is explainable in terms of this aspect. Thus, we calculate a hit-ratio for the test set (i.e., the number of successful hits divided by the number of samples). A large hit-ratio indicates the compressed model generates the predictions based on the pivotal aspect with larger attention wights demonstrating the model achieves better explainability. 

\begin{figure}[ht]
	\centering 
	\includegraphics[width=0.80\linewidth]{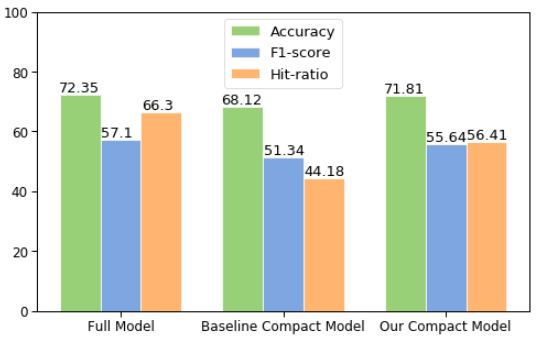}
	\caption{Evaluation of model explainability on Rest-2014.}
	\label{fig:rest}
\end{figure}

As shown in Figure \ref{fig:rest} and Table \ref{baseline}, our compressed model achieves a comparable performance with the full model in both clean and adversarial sample settings. In addition, compared to the conventionally trained compressed models (Figure \ref{fig:rest}), our model achieves a better performance in terms of both accuracy and explainability (hit-ratio +12\%). This improvement is mainly due to the fact that the features of our compressed model are derived from the full model via an interpretable feature mapping guided by the pivotal aspects.    

\subsection{Model Personalization}

\begin{figure}[hbt!]
	\centering 
	\includegraphics[width=0.90\linewidth]{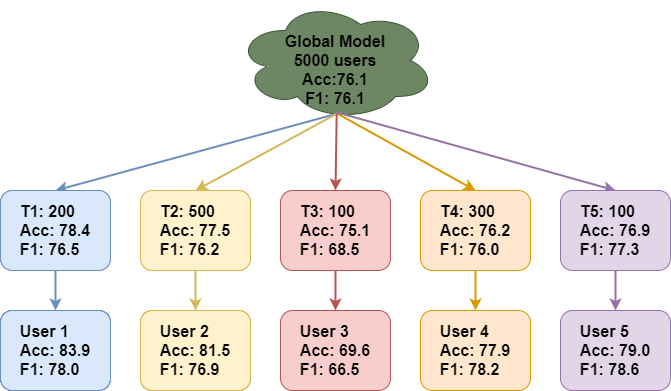}
	\caption{On-device personalization on Senti140.}
	\label{fig:twitter}
\end{figure}

As it is hard to get a non-trivial number of data samples on users' devices to train a personal model, we propose a two-step strategy to retain the model personality. We first train a global model on a global dataset collected from 5000 users. Then we fine-tune it on each user's private training data while deploying it on-device. Additionally, we find it effective to leverage the hyper-parameters $T$ in the label distillation layer while training the global model to achieve better performance on the local data. As shown in Figure \ref{fig:twitter}, each on-device model achieves its best performance with a different value of $T$. In more details, in devices such as User 1, the fine-tuned personal model achieves a better model performance than the global model, probably due to the higher quality of the private training data. Otherwise, we use the global model instead of the fine-tuned personal model (e.g. User 3) and get improved accuracy (69.6$\rightarrow$75.1) and F1 score (66.5$\rightarrow$68.5). 

% In the first step, we first train a global model with the global dataset collected from 5,000 users, then leverage the hyper-parameters $T$ in the label distillation layer to fine tune a personalized compressed model for each user (client) achieving best testing performance. In the second step, we fine tune the compressed model while deploying it on-device with each users' private training data. As shown in Figure \ref{fig:twitter}, each on-device model achieves its best performance on a different value of $T$, demonstrating the effect of model personalization. In devices such as User 1, the fine tuned model achieves a higher model performance than the original compressed model before personalization, probably due to the higher quality of the private training data. Otherwise, we use the original compressed model instead of the fine tuned model (e.g. User 3) and get improved accuracy (69.64$\rightarrow$75.15) and F1 score (66.56$\rightarrow$68.57). 

%\subsection{On-device Performance}
%(This part will report the performance of the models deployed on mobile devices.)

\subsection{Model Size Selection and Ablation Study}

In Table \ref{influence}, we investigate the effect of embedding dimension ($d$) and latent feature size ($H$) on the model compression performance. We set both $d$ and $H$ values at 100 for our full model, resulting in the model size of 2 million (2M) parameters. Whereas for the three compressed models, we set both $d$ and $H$ values at 5, 10, or 20, resulting in the model sizes of 100K, 200K, and 400K parameters, respectively. The compressed models with larger $d$ and $H$ achieve better performance due to lower information loss. In Table \ref{ablation}, we show an ablation study to examine the effectiveness of the key components (e.g., embedding feature mapping layer, latent feature mapping layer, and label distillation layer) using three different datasets. The compressed models with all the components achieve the best performance in both clean and adversarial sample accuracy and macro-F1 score. The importance of a layer component increases as it gets closer to the output.

\begin{table}[hbt!]
	\centering
	\resizebox{0.48\textwidth}{!}{
		\begin{tabular}{l|c|c|c|c|c}
			\hline
			\multirow{2}*{\textbf{Dataset}}  & \multirow{2}*{\textbf{Dimension}} & \textbf{Clean}        		& \textbf{Replaceone}   & \textbf{Random}  & \textbf{Gradient}     \\
			&& Acc/F1          				& AdvAcc/AdvF1      	& AdvAcc/AdvF1         & AdvAcc/AdvF1          \\ 
			\hline
			\multirow{3}*{Yelp} &5    & 53.7/37.3            	    	& 37.2/26.3      	& 52.1/35.8		    & 52.0/36.1   			  \\
			&10       				  & 61.6/46.7           	  		& 45.3/\textbf{32.2}  & 58.4/46.6  	 & 62.5/48.0    			  \\
			&20        				  & \textbf{61.8}/\textbf{46.8}   & \textbf{47.3}/31.4  & \textbf{61.3/48.6} & \textbf{65.9/51.2}   \\ \hline
			\hline
			\multirow{3}*{Yahoo} &5   					  & 50.1/40.6            	  		& 35.8/28.8       	& 46.8/38.0  		   & 41.6/33.6     			  \\
			&10   					  & 71.5/65.1           	  		& 49.6/48.3       	& 71.1/65.4 		& 66.6/64.6     			  \\
			&20						  & \textbf{72.2}/\textbf{65.8}   & \textbf{51.9/49.3}  & \textbf{71.2/67.3} & \textbf{69.0/66.0}     \\ \hline
			\hline
			\multirow{3}*{AG's News}  &5    				  & 90.3/67.8           	  		& 68.1/50.8       	& 89.4/66.9 		   & 88.1/65.8     			  \\
			&10     				  & \textbf{91.1}/\textbf{68.4}   & 79.6/62.0       	& \textbf{92.1/69.0} & \textbf{90.2/67.4}   \\
			&20   				      & 90.7/68.0             	 	& \textbf{79.9/62.5}  & 89.6/67.1  		   & 89.6/66.5     			  \\ \hline
		\end{tabular}
	}
	\caption{Comparison of diverse sizes of compressed models.} %5, 10, and 20 represent the embedding dimension ($d$) and latent feature size ($H$).}	
	\label{influence} 
\end{table}

\begin{table}[hbt!]
	\centering
	\resizebox{0.38\textwidth}{!}{
		\begin{threeparttable}
			\begin{tabular}{l|l|c|c}
				\hline
				\multirow{2}*{\textbf{Dataset}}     & \multirow{2}*{\textbf{Layers}}   & \textbf{Clean} & \textbf{Replaceone}        \\ 
				& 	 			    & Acc/F1 	    & AdvAcc/AdvF1       \\ 
				\hline
				\multirow{4}*{Yelp}         & w/o Embedding & 60.3/47.3    	& 37.4/26.1\\ 
				& w/o Latent       & 60.7/47.5    	& 38.6/28.0\\ 
				& w/o Label        & 61.0/47.4    	& 36.8/27.2\\ 
				& All              & \textbf{61.6/47.7} & \textbf{45.3/32.2}\\ \hline
				\hline
				\multirow{4}*{Yahoo}        & w/o Embedding & 70.0/63.4     & 45.2/41.9\\
				& w/o Latent        & 70.4/64.1     & 45.3/42.0\\ 
				& w/o Label         & 70.9/64.4     & 44.8/40.6\\ 
				& All               & \textbf{71.5/65.1}  & \textbf{49.6/48.3} \\ \hline
				\hline
				\multirow{4}*{AG's News}    & w/o Embedding & 89.8/66.8     & 68.5/51.4 \\ 
				& w/o Latent            & 90.0/67.2     & 69.6/51.9\\ 
				& w/o Label             & 90.4/67.9     & 75.7/56.2\\ 
				& All                   & \textbf{91.1/68.4} & \textbf{79.6/62.0}\\ \hline
			\end{tabular}
			%\begin{tablenotes}
			%\item[1] w/o - without.
			%\end{tablenotes}
		\end{threeparttable}
	}
	\caption{Contribution of different layers of our model.} %Replaceone represents an adversarial attack method.}
	\label{ablation} 
\end{table}

\section{Conclusions and Future Work}

In this work, we design a new training scheme for model compression ensuring adversarial robustness, explainability, and personalization for NLP applications. Our novel aspect-based feature mapping and label distillation minimize the information loss and maximize the explainability of model compression, and the new training objective ensures model's adversarial robustness. The performance can be further improved via on-device personalization. Unlike the adversarial training that is computationally prohibitive and limited to the known attacks, our training scheme is sufficiently flexible, effective, and robust for on-device NLP applications. In future work, we will generalize our training scheme to other on-device NLP tasks, such as Question Answering, Reading Comprehension, and Neural Machine Translation. Furthermore, transformer-based models (e.g BERT \cite{devlin2018bert}) gain a lot of attention dealing with NLP tasks recently. There are several recent studies on compressing BERT models, e.g., MobileBERT \cite{sun2020mobilebert} and TinyBERT \cite{jiao2019tinybert}. These approaches mainly focus on model compression, while none of them deals with model adversarial robustness and explainability. We will extend our training scheme to the widely used transformer-based models.

\clearpage
\bibliographystyle{IEEEtran}
\bibliography{IEEEabrv,cite}

\end{document}